\documentclass[11pt,a4paper]{article}
\usepackage[hyperref]{emnlp2020}
\usepackage{times}
\usepackage{latexsym}

\usepackage{microtype}

\usepackage{color}
\usepackage{times}
\usepackage{soul}
\usepackage{url}
\usepackage[utf8]{inputenc}
\usepackage{graphicx}
\usepackage{amsmath}
\usepackage{amssymb}
\usepackage{amsthm}
\usepackage{booktabs}
\usepackage{algorithm}
\usepackage{algorithmic}
\usepackage{multirow}
\usepackage{tikz}
\usepackage{xcolor}
\usepackage{pifont}

\aclfinalcopy 

\newcommand*\colourcheck[1]{%
  \expandafter\newcommand\csname #1check\endcsname{\textcolor{#1}{\large \ding{52}}}%
}
\colourcheck{green}

\newcommand*\colourcross[1]{%
  \expandafter\newcommand\csname #1cross\endcsname{\textcolor{#1}{\large \ding{56}}}%
}
\colourcross{red}

\title{Neural Retrieval for Question Answering with\\ Cross-Attention Supervised Data Augmentation} 

\author{
Yinfei Yang\textsuperscript{$a$},
Ning Jin\textsuperscript{$b$},
Kuo Lin\textsuperscript{$b$},
Mandy Guo\textsuperscript{$a$},
Daniel Cer\textsuperscript{$a$} \AND
  {\rm\textsuperscript{$a$}Google Research}\\Mountain View, CA, USA \And
  {\rm\textsuperscript{$b$}Google Cloud AI}\\Sunnyvale, CA, USA 
}
\date{}

\begin{document}
\maketitle

\begin{abstract}
Neural models that independently project questions and answers into a shared embedding space allow for efficient continuous space retrieval from large corpora. Independently computing embeddings for questions and answers results in late fusion of information related to matching questions to their answers. While critical for efficient retrieval, late fusion underperforms models that make use of early fusion (e.g., a BERT based classifier with cross-attention between question-answer pairs). We present a supervised data mining method using an accurate early fusion model to improve the training of an efficient late fusion retrieval model. We first train an accurate classification model with cross-attention between questions and answers. The accurate cross-attention model is then used to annotate additional passages in order to generate weighted training examples for a neural retrieval model. The resulting retrieval model with additional data significantly outperforms retrieval models directly trained with gold annotations on Precision at $N$ (P@N) and Mean Reciprocal Rank (MRR).

\end{abstract}

\section{Introduction}
Open domain question answering~(QA) involves finding answers to questions from an open corpus~\cite{surdeanu-etal-2008-learning,yang-etal-2015-wikiqa,chen-etal-2017-reading,reqa}. The task has led to a growing interest in scalable end-to-end retrieval systems for question answering. Recent neural retrieval models have shown rapid improvements, surpassing traditional information retrieval~(IR) methods such as BM25~\cite{reqa,lee2019latent,karpukhin2020dense}. 

When QA is formulated as a reading comprehension task, cross-attention models like BERT~\cite{devlin-etal-2019-bert} have achieved \textit{better-than-human} performance on benchmarks such as the Stanford Question Answering Dataset (SQuAD)~\cite{rajpurkar-etal-2016-squad}.
Cross-attention models are especially well suited for problems involving comparisons between paired textual inputs, as they provide \textit{early} fusion of fine-grained information within the pair. This encourages careful comparison and integration of details across and within the two texts. 

However, early fusion across questions and answers is a poor fit for retrieval, since it prevents pre-computation of the answer representations. Rather, neural retrieval models independently compute embeddings for questions and answers typically using dual encoders for fast scalable search~\cite{henderson2017efficient,guo-etal-2018-effective,e2eretrieval,yang2019multilingual}. Using dual encoders results in \textit{late} fusion within a shared embedding space.

For machine reading, early fusion using cross-attention introduces an inductive bias to compare fine grained text spans within questions and answers. This inductive bias is missing from the single dot-product based scoring operation of dual encoder retrieval models. Without an equivalent inductive bias, late fusion is expected to require additional training data to learn the necessary representations for fine grained comparisons.

To support learning improved representations for retrieval, we explore a supervised data augmentation approach leveraging a complex classification model with cross-attention between question-answer pairs. 
Given gold question passage pairs, we first train a cross-attention classification model as the supervisor.
Then any collection of questions can be used to mine potential question passage pairs under the supervision of the cross-attention model.
The retrieval model training benefits from additional training pairs annotated with the graded predictions from the cross-attention model augmenting, the existing gold data. 

Experiments are reported on MultiReQA-SQuAD and MultiReQA-NQ, with retrieval models establishing significant improvements on Precision at $N$ (P@N) and Mean Reciprocal Rank (MRR) metrics.

\section{Neural Passage Retrieval for Open Domain Question Answering}

Open domain question answering is the problem of answering a question from a large collection of documents~\citep{trec,chen-etal-2017-reading}.
Systems usually follow a two-step approach: first retrieve question relevant passages, and then scan the returned text to identify the answer span using a reading comprehension model~\citep{Jurafsky:2009:SLP:1214993,kratzwald-feuerriegel-2018-adaptive,yang-etal-2019-end}.
Prior work has focused on the answer span annotation task and has even achieved super human performance on some datasets. However, the evaluations implicitly assume the trivial availability of passages for each question that are likely to contain the correct answer. While the retrieval task can be approached using traditional keyword based retrieval methods such as BM25, there is a growing interest in developing more sophisticated neural retrieval methods~\cite{lee2019latent,guu2020realm,karpukhin2020dense}.

\section{Retrieval Question-Answering~(ReQA)}
\label{sec:reqa}

\citet{reqa} introduced the Retrieval Question-Answering~(ReQA) task that has been rapidly adopted by the community~\cite{multireqa,Chang2020Pre-training,ma2020zeroshot,zhao2020talk,lareqa}. Given a question, the task is to retrieve the answer sentence from a corpus of candidates. ReQA provides direct evaluation of retrieval, independent of span annotation.
Compare to Open Domain QA, ReQA focuses on evaluating the retrieval component and, by construction, avoids the need for span annotation.

We explore the proposed approach on the MultiReQA-NQ and MultiReQA-SQuAD tasks.\footnote{\url{https://github.com/google-research-datasets/MultiReQA}}
MultiReQA~\cite{multireqa} established standardized training / dev / test splits.
Statistics for the MultiReQA-NQ and MultiReQA-SQuAD tasks are listed in Table \ref{tab:reqa_stats}.

\begin{figure}[!t]
  \centering
  \includegraphics[width=0.48\textwidth]{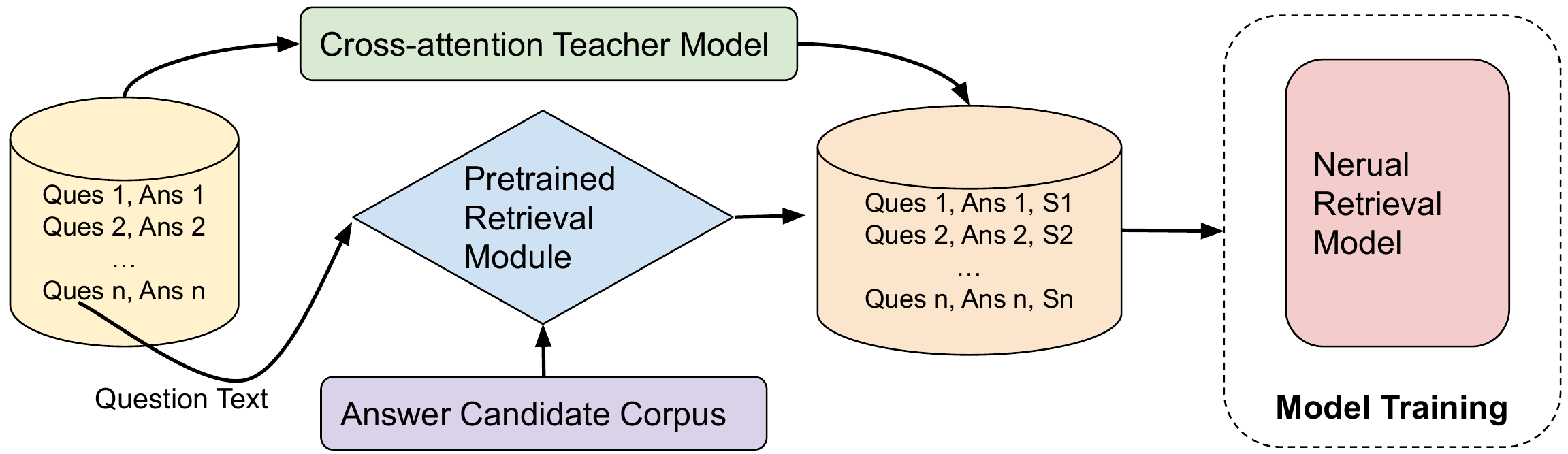}
  \caption{
    Use of a cross-attention model for the supervised mining of additional QA pairs. Our accurate cross-attention model supervises the mining process by identifying new previously unannotated positive pairs.  
    Mined QA pairs augment the original training data for the dual encoder based neural passage retrieval model.
  }
  \label{fig:pipeline}
\end{figure} 

\begin{table}[htb]
    \small
    \centering
    \begin{tabular}{ l | r | r r  } \hline
    \multirow{2}{*}{\textbf{Dataset}} & \multirow{2}{*}{\textbf{Training Pairs}} & \multicolumn{2}{c}{\textbf{Test}}  \\ 
    & & Questions & Candidates  \\ \hline
    NQ            & 106,521 & 4,131 & 22,118 \\
    SQuAD         & 87,133 & 10,485 & 10,642 \\
    \hline
    \end{tabular}
    \caption{Statistics of MutiReQA NQ and SQuAD tasks: \# of training pairs, \# of questions, \# of candidates.}
    \label{tab:reqa_stats}
\end{table}

\section{Methodology}

Figure \ref{fig:pipeline} illustrates our approach using a cross-attention classifier to supervise the data augmentation process for training a retrieval model. After training the cross-attention model, we retrieve additional potential answers to questions in the training set using an off-the-shelf retrieval system\footnote{Note the approach can also be applied to any collection of questions, even for those without ground truth answers.}. The predicted scores from the classification model are then used to weight and filter the retrieved candidates with positive examples serving as weighted silver training data for the retrieval model.

\subsection{BERT Classification Model}
\label{sec:cqnli}
Cross-attention models like BERT are often used for re-ranking after retrieval and can significantly improve performance as they allow for fine-grained interactions between paired inputs~\cite{nogueira2019multistage,han2020learningtorank}.
Here we formalize a binary classification task for predicting the question answer relatedness.
We use the question-answer pairs from the training set as our positive examples. Negatives are sampled for each question using following strategies with 1:1:1 ratio:
\begin{itemize}
    \item[1]  For each question, we sample a sentence from the 10 nearest neighbors returned by a term based BM25~\cite{Robertson:2009} from a sentence pool containing all supporting documents in a corpus. Sampled sentences are paired with questions as negative examples.
    
    \item[2]  Similar to our BM25 negatives and drawing from the same sentence pool, we sample the 10 nearest neighbors using the Universal Sentence Encoder - QA (USE-QA)~\cite{yang2019multilingual}. Sampled sentences are paired with the question that selected it and labeled as negative. 
    
    \item[3]  Each question is paired with a sentence randomly sampled from its supporting documents, excluding the question's gold answer.
\end{itemize}
A BERT model is fine-tuned following the default setup from the \citet{devlin-etal-2019-bert}.

\subsection{Dual-Encoder Retrieval Model}

\begin{figure}[!htb]
  \centering
  \includegraphics[width=0.48\textwidth]{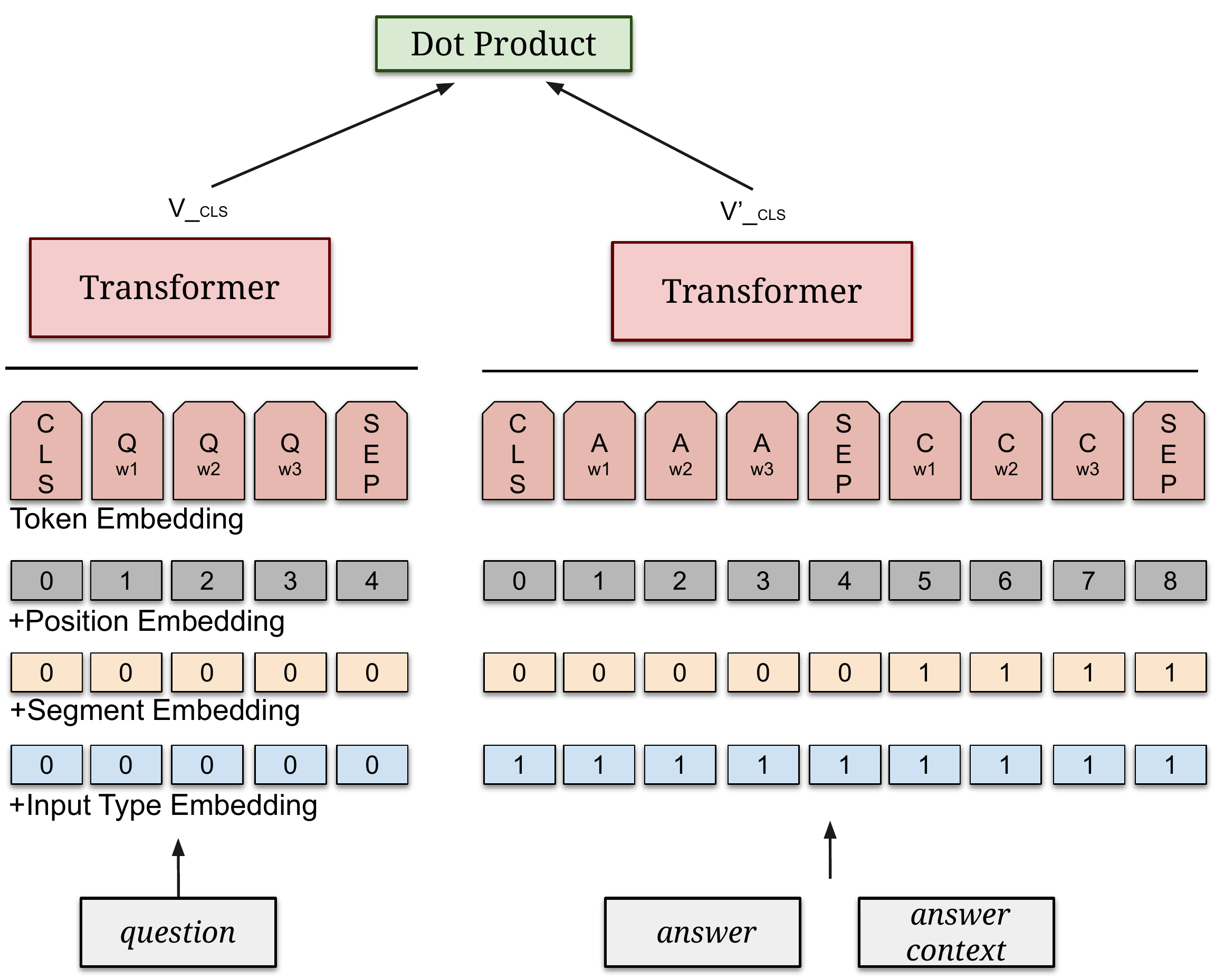}
  \caption{
    The BERT dual encoder architecture. The answer and context are concatenated and fed into the answer encoder. Figure from \cite{multireqa}.
  }
  \label{fig:bert_dual_encoder}
\end{figure} 

We follow \citet{multireqa} and employ a BERT based dual-encoder model for retrieval.
The dual encoder model critically differs from the cross-attention model in that there is no early interactions~(cross-attention) between the question and answer. The resulting independent encodings are only combined in the final dot-product scoring a pair.
The same BERT encoder is used for questions and answers with the output of the CLS token taken as the output encoding.
For answers, the answer and context are concatenated and segmented using the segment IDs from the original BERT model.
A learned \textit{input type embedding} is added to each token embeddings to distinguish questions and answers within the encoding model.


\subsection{Mining Augmented Training Pairs}

We create an augmented training set for the retrieval model using the cross-attention model. For each question in the training set, we use USE-QA to mine the top 10 nearest neighbors from the entire training set, and then remove those retrieved pairs which are true positives. 
Next the cross-attention model is used to score the retrieved pairs. 
The neural retrieval model is trained on the combination the scored pairs and the original question-answer pairs from training set. The original pairs are assigned a score 1.

\subsection{Weighted In-batch Softmax for Dual-Encoder Retrieval Model}

The neural retrieval model is trained using the batch negative sampled softmax loss~\cite{e2eretrieval} in equation \ref{eq:bayes_weighted}. We modify the standard formulation to include a weight, $w(x,y)$, for each pair.
\begin{equation}
\label{eq:bayes_weighted}
\mathcal{J'} = \sum_{(x, y) \in Batch} w(x,y) \frac{e^{\phi(x, y)}} {\sum_{\bar{y} \in \mathcal{Y}} e^{\phi(x, \bar{y})}}
\end{equation}
 
The $w(x,y)$ is set to 1 if $(x, y)$ is the ground truth positive pair and $p(x, y)^2$ otherwise, where $p(x, y)$ is the probability output from the cross-attention model if it is not a ground truth example.

\section{Evaluation}

In this section we evaluate the proposed approach using the MultiReQA evaluation splits for NQ and SQuAD. 
Models are assessed using Precision at N (P@N) and Mean Reciprocal Rank (MRR). Following the ReQA setup~\cite{reqa}, we report P@N for N=$[1,5,10]$.
The P@N score tests whether the true answer sentence appears as the top-N-ranked candidates.
And MRR is calculated as $\textrm{MRR} = \frac{1}{N} \sum_{i=1}^{N} \frac{1}{\textit{rank}_i}$, where $N$ is the total number of questions, and \emph{rank}$_i$ is the rank of the first correct answer for the $i$th question. 

\subsection{Configurations}

We fine-tune the public English BERT cross-attention models using Batch size of 256; weighted Adam optimizer with learning rate 3e-5. 
Each model is fine-tuned for 10 epochs.
We experiment with both Base and Large BERT models.
All hyper-parameters are set using a development set splitted out from the training data~(10\%).
While mining for silver data, we keep only the mined examples with cross-attention model scores predicted as positive (with score $\geq0.5$).

The BERT Base model is used to initialize the dual encoder retrieval model. 
During training we use a batch size of 64, and a weighted Adam optimizer with learning rate 1e-4.
The maximum input length is set to 96 for questions and 384 for answers.  
Models are trained for 200 epochs.  Following \citet{multireqa}, we use the BERT CLS token as the text embedding for retrieval.
The embeddings are $l_2$ normalized.
Hyper-parameters are manually tuned on a held out development set.

\begin{table}[!t]
\small
    \centering
    \begin{tabular}{l | r r | r r } \hline
    \multirow{2}{*}{\textbf{Models}}& \multicolumn{2}{c|}{\textbf{NQ}} &\multicolumn{2}{c}{\textbf{SQuAD}} \\
    & ACC & AUC-PR & ACC & AUC-PR \\ \hline
    \rule{-2pt}{8pt}
    Majority                    & 73.7 & -- & 74.8 & -- \\
    BERT\textsubscript{dual\_encoder}   & 75.8 & 49.3 & 80.3 & 62.0 \\ 
    (S)~BERT\textsubscript{Base}        & 84.3 & 92.8 & 92.6 & 96.5 \\
    (S)~BERT\textsubscript{Large}       & 84.9 & 93.5 & 93.6 & 97.1 \\ \hline
    \end{tabular}
    \caption{Accuracy (ACC) and  area under the precision-recall curve (AUC-PR) for the classification task.  \textbf{Majoirty} is a simple baseline that always predict false. \textbf{(T)} indicates the model is a supervisor model candidate.}
    \label{tab:cqnli_result}
\end{table}

\begin{table*}[!t]
    \centering
    \begin{tabular}{l | r r r r| r r r r} \hline
    \multirow{2}{*}{\textbf{Models}}& \multicolumn{4}{c|}{\textbf{NQ}} &\multicolumn{4}{c}{\textbf{SQuAD}} \\
    & P@1 & P@5 & P@10 & MRR & P@1 & P@5 & P@10 & MRR \\ \hline
    \rule{-2pt}{8pt}
    BM25                                            & 24.7 & -- & -- & 36.6 & 62.8 & -- & -- & 70.5 \\
    USE-QA                                          & 24.7 & -- & -- & 34.7 & 51.0 & -- & -- & 62.1 \\ 
    BERT\textsubscript{dual\_encoder}               & 44.7 & 77.1 & 85.1 & 58.9 & 62.8 & 85.4 & 91.0 & 72.8 \\
    BERT\textsubscript{dual\_encoder}~Augmented     & 53.3 & 82.3 & 88.5 & 65.9 & 63.8 & 86.1 & 91.6 & 73.7 
    \end{tabular}
    \caption{Precision at N(P@N) (\%) N=$[1,5,10]$ and Mean Reciprocal Rank (MRR) (\%) on the MultiReQA tasks.}
    \label{tab:reqa_result}
\end{table*}

\subsection{Performance for the Classification Task}

The classification data created using the method from section \ref{sec:cqnli} contains a total of 531k and 469k training examples for NQ and SQuAD, respectively.
Test sets extracted from the SQuAD and NQ test splits contain 15k and 41k examples.\footnote{The positive~/~negative ratio is roughly 1:3.}

Table \ref{tab:cqnli_result} shows the performance of the cross-attention models. We compare performance to the majority baseline which always predict false and the BERT\textsubscript{dual\_encoder} retrieval model which uses cosine similarity for prediction.
BERT based models outperform the baselines by a wide margin,\footnote{The poor performance of BERT\textsubscript{dual\_encoder} is also aligned with the hypothesis that cosine similarity score is not a globally consistent measurement of how good a pair~\cite{guo-etal-2018-effective}.} 
with the BERT large model achieving the highest performance on all metrics.
This is consistent with our hypothesis that late fusion  models outperforms the retrieval model on the task.
Both models achieve better performance on the SQuAD than NQ. 
The SQuAD task has higher token overlap, as described in \ref{sec:reqa}, making the task somewhat easier.
We use the \textbf{BERT large model} to supervise the data augmentation step in the next section.

\subsection{Mined Examples}
We mined the SQuAD and NQ training data to construct additional QA pairs. 
After collecting and scoring addition pairs using the method described in section 4.3, 
we obtained 53\%~(56,148) and 12\%~(10,198) more examples for NQ and SQuAD, respectively. Table \ref{tab:mined_examples} illustrated the examples retrieved by USE-QA and predicted as positive examples by the classification model. 
Both of the examples are clear positives.

Much less data is mined for SQuAD then NQ. 
We believe it is because of the way SQuAD was created, whereby workers write the questions based on the content of a particular article.
The resulting questions are much more specific and biased toward a particular question types, for example \citet{reqa} shows almost half of the SQuAD questions are \textit{what} questions.
Another reason is that the candidate pool for SQuAD is only half that of NQ, resulting in questions having fewer opportunities to be matched to good additional answers.

\begin{table}[!h]
\resizebox{\linewidth}{!}{%
\begin{tabular}{c|p{6cm}}
\textbf{Score} & \multicolumn{1}{c}{\bf Silver QA Pair} \\
\hline
\multirow{5}{*}{0.92} & \textbf{Q:} what are the names of the two old muppets in the balcony that heckle everyone ? \\ 
& \textbf{A:} Statler and Waldorf are a pair of Muppet characters known for their cantankerous opinions and shared penchant for heckling. \\ \hline
\multirow{4}{*}{0.90} & \textbf{Q:} where the phrase dressed to the nines come from \\
& \textbf{A:} It appears in book six of Jean - Jacques Rousseau 's Confessions , his autobiography ... \\
\end{tabular}}
\caption{Scored examples from cross-attention classification model.}
\label{tab:mined_examples}
\end{table}

\subsection{Results on the Retrieval QA}

Table \ref{tab:reqa_result} shows the P@N and MRR@100 of the retrieval models on MultiReQA-SQuAD and MultiReQA-NQ.
The first two rows show the result from two simple baselines BM25~\cite{Robertson:2009} and USE-QA reported from \citet{multireqa}.
BM25 remains as a strong baseline, especially with 62.8\%~P@1 and 70.5\%~MRR for SQuAD.
The performance on NQ is much lower, as there is much less token overlap between NQ questions and answers.
USE-QA matches the performance of BM25 on NQ but performs worse on SQuAD.\footnote{We note that USE-QA can be fine-tuned using the training set, which will usually significantly outperform the default USE-QA model as demonstrated in \citet{multireqa}.
}

BERT\textsubscript{dual\_encoder} trained on the NQ and SQuAD training set performs very strong compared to the baselines, especially on NQ with a +20 point improvement on NQ.~\footnote{Our Bert\textsubscript{dual\_encoder} performs much better than the one reported in \citet{multireqa}, we found simply train the model longer significantly improves the model performance.}
Our P@1 on SQuAD matches BM25, but we achieve an MRR that is +2.3 points better.
Performance is further improved by including the augmented data from our cross-attention model, obtaining 53.3\% P@1 and 65.9\% MRR on NQ, which is an +8.6\% and a +7.0\% improvement on P@1 and MRR, respectively, comparing with the second best model.

Compare to NQ, the improvement on SQuAD is rather marginal.
The augmented BERT\textsubscript{dual\_encoder} retrieval model only achieves slightly improved performance on SQuAD, with +1 points for both of P@1 and MRR.
As discussed in last section, we mine much less data on SQuAD compare with NQ, with only ~10\% more data on top of the original training set.
As demonstrated by the strong BM25 performance and shown in \cite{multireqa},
the SQuAD question answer pairs have higher token overlap between question and answers, eliminating the advantage of the neural methods to implicitly model more complex semantic relationships.

\section{Conclusion}
In this paper, we propose a novel approach for making use of an early fusion classification model to improve late fusion retrieval models.
The early fusion model is used to supervised data mining that augments the training data for the later model.
The proposed approach mines 53\%~(56,148) and 12\%~(10,198) more examples for MultiRQA-NQ and MultiRQA-SQuAD, respectively.
The resulting retrieval models improve +8.6\% and +1.0\% on P@1 on NQ and SQuAD, respectively.
The current pipeline assumes there exists annotated in-domain question answer pairs to train the cross-attention model.
With a strong general purpose cross-attention model, our supervised data mining method could be modified to train in-domain retrieval models without gold question answer pairs. 
We leave this direction to the future work.

\bibliographystyle{acl_natbib}
\bibliography{anthology,emnlp2020}

\end{document}